\documentclass[letterpaper, 10 pt, conference]{ieeeconf}  

\IEEEoverridecommandlockouts                              

\usepackage{amssymb}  
\usepackage{times}
\usepackage{soul}
\usepackage{url}
\usepackage{xcolor}

\makeatletter
\let\NAT@parse\undefined
\makeatother

\usepackage{hyperref}
\hypersetup{
    colorlinks,
    linkcolor={red!50!black},
    citecolor={blue!50!black},
    urlcolor={blue!80!black}
}
\usepackage[utf8]{inputenc}
\usepackage[small]{caption}
\usepackage{graphicx}
\usepackage{algorithm}
\usepackage{algpseudocode}
\usepackage{booktabs}
\usepackage{amsmath}
\DeclareMathOperator*{\argmax}{arg\,max}
\usepackage{amsfonts}
\usepackage{mathtools}
\usepackage{verbatim}
\usepackage{sidecap}

\title{\LARGE \bf
Reasoning about Counterfactuals to Improve\\ Human Inverse Reinforcement Learning
}

\author{Michael S. Lee, Henny Admoni, and Reid Simmons
\thanks{The authors are with the Robotics Institute at Carnegie Mellon University
        {\tt\small{ml5, hadmoni, rsimmons@}andrew.cmu.edu}. This work was supported by the Office of Naval Research award N00014-18-1-2503.}. %
}

\begin{document}

\maketitle
\thispagestyle{empty}
\pagestyle{empty}

\begin{abstract}

To collaborate well with robots, we must be able to understand their decision making. Humans naturally infer other agents' beliefs and desires by reasoning about their observable behavior in a way that resembles inverse reinforcement learning (IRL). Thus, robots can convey their beliefs and desires by providing demonstrations that are informative for a human learner's IRL. An informative demonstration is one that differs strongly from the learner's expectations of what the robot will do given their current understanding of the robot's decision making. However, standard IRL does not model the learner's existing expectations, and thus cannot do this counterfactual reasoning. We propose to incorporate the learner's current understanding of the robot's decision making into our model of human IRL, so that a robot can select demonstrations that maximize the human's understanding. We also propose a novel measure for estimating the difficulty for a human to predict instances of a robot's behavior in unseen environments. A user study finds that our test difficulty measure correlates well with human performance and confidence. Interestingly, considering human beliefs and counterfactuals when selecting demonstrations decreases human performance on easy tests, but increases performance on difficult tests, providing insight on how to best utilize such models.

\end{abstract}

\section{INTRODUCTION}

\noindent Our capacity to deploy, collaborate, and co-exist fluently with robots is contingent on our ability to understand their decision-making. For example, an engineer certifying the navigation policy of a ground delivery robot may ask, ``Does the robot understand all the terrains it might encounter well enough to successfully balance efficiency and safety?'' Moreover, new owners of an autonomous vacuum gauging how much of their floor to declutter may wonder, ``How much clutter will the robot tolerate before it steers clear of an area to ensure it does not get stuck?'' More generally, various human stakeholders, from manufacturers, to co-workers, to owners, must be able to reliably predict robot behavior for robots to be seamlessly be integrated into society. Toward this end, we explore how robots can convey their decision-making and subsequent policies to human learners.

One important way that humans communicate and comprehend each others' decision-making is through demonstrations. Human behavior is commonly modeled as being driven by reward functions \cite{jara2016naive}, which can be inferred through reasoning akin to inverse reinforcement learning (IRL) \cite{baker2009action,jara2019theory}. Once someone's reward function is known, humans may then (approximately) deduce their corresponding reward-maximizing behavior through planning \cite{shteingart2014reinforcement}. Furthermore, although a robot could convey its reward function to a human directly, we posit that it can be nontrivial for humans to precisely derive the optimal behavior from a numerical reward function, particularly if there a is large number of reward features or the reward features themselves are hard to define (e.g. aggressiveness for driving). Indeed, a study by Sukkerd et al. \cite{sukkerd2020tradeoff} that suggests that humans better understand a robot policy if demonstrations accompany a numerical reward function. Motivated by how humans communicate and comprehend behavior, we use demonstrations informative for IRL to convey decision-making.



\begin{figure*}[!t]
\centering
\includegraphics[width=\textwidth]{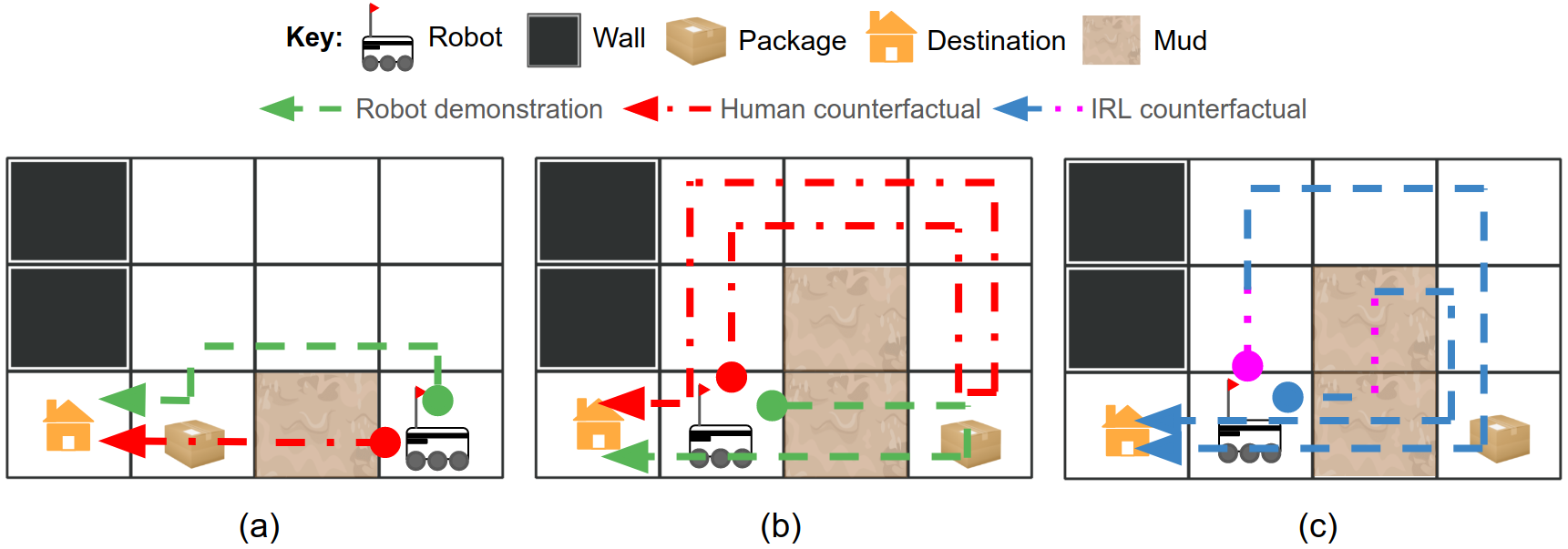} 
\caption{\textbf{(a)} A robot's optimal demonstration (green) is shown in contrast to a suboptimal counterfactual alternative (red). \textbf{(b)} A robot's optimal demonstration is shown in contrast to a counterfactual likely considered by a human who has seen the demonstration in (a). \textbf{(c)} Sample counterfactual alternatives to the robot's trajectory in (b) considered by standard IRL --- generated by deviating from the robot's path by one action (pink), then following the robot's optimal policy afterward (blue). Note that neither matches the human's counterfactual.}
\label{fig:combined}
\vspace{-3mm}
\end{figure*}

For IRL, the information a demonstration reveals regarding the underlying reward function inherently depends on the \textit{counterfactuals} (i.e. alternative, suboptimal demonstrations) that are considered. Picture a delivery robot that must bring a package to the destination, whose reward function balances traveling through difficult terrain, like mud, and the overall number of actions it takes (i.e. steps). To convey its reward function, imagine the robot provides a human with the demonstration in Fig. \ref{fig:combined}a. Because the robot takes a two-action detour to avoid the mud instead of going through it (a natural counterfactual), the human may infer that the robot associates a negative reward with going through mud.

After providing this first demonstration, the robot considers what to demonstrate next to convey more information regarding its reward function. Importantly it knows that the human likely knows that mud is costly from the first demonstration, but does not know \emph{how} costly. For instance, the human may reasonably believe that the robot would take a four-action detour when faced with two mud patches (Fig. \ref{fig:combined}b). However, the robot knows that its ratio of mud to action reward is -3 to -1 and that consequently, it would simply go through the mud in Fig. \ref{fig:combined}b to maximize its reward. Seeing how its direct path meaningfully differs from the human's likely counterfactual that detours heavily, the robot considers this to be a very informative demonstration to provide next to upper-bound the human's belief of the cost of mud.

Standard IRL \cite{ng2000algorithms}, however, does not model the learner's beliefs and would fail to consider this detouring human counterfactual. Instead, standard IRL (henceforth referred to as IRL for the sake of brevity) enumerates all trajectories that differ by a single initial action as counterfactuals. Two sample IRL counterfactuals are shown in Fig. \ref{fig:combined}c, but neither matches the intuitive human counterfactual of completely detouring around the mud. As a result, IRL has the potential to both under- and over-estimate the informativeness of a demonstration to a human by considering the wrong counterfactuals or considering too many, respectively.


This work aims to more effectively convey robot decision-making to humans by using a model of human beliefs to reason about demonstrations. First, we evaluate the informativeness of demonstrations based on counterfactual trajectories likely to be considered by the human rather than those generated via one-action deviations by standard IRL. Second, we aim to further improve demonstration selection by incrementally increasing (and scaffolding) the number of unique reward features that are conveyed (e.g. the delivery domain not only has mud, but also a battery recharge station). And toward testing a learner's understanding of a robot's decision-making, we also propose a novel measure for estimating the difficulty for a human to predict instances of a robot's behavior in unseen environments.


Our user study finds that our test difficulty measure correlates well with human performance and confidence but finds no effect for feature scaffolding. Considering human beliefs on robot decision making in selecting informative demonstrations decreases human performance on easy tests, but increases performance for difficult tests, providing insight on how to best utilize such human models.

\section{Related Work}

\textbf{Policy Summarization:} Policy summarization aims to convey a global understanding of a robot's policy to a user through select states and actions \cite{Amir2019}. The first approach relies on heuristics such as entropy or differences in Q-values to select states and actions to show \cite{huang2018establishing,amir2018highlights}.

We instead build on the second approach is based on  machine teaching \cite{zhu2018overview}. Machine teaching aims to teach a target model (e.g. reward function) to a learner with a given learning model (e.g. IRL) using a minimal set of teaching examples (e.g. demonstrations). Methods for conveying a robot's reward function and/or behavior to humans are surveyed by Sanneman et al. \cite{sanneman2022empirical} and Booth et al. \cite{booth2022revisiting} and we summarize a few relevant works below.

Brown and Niekum \cite{brown2019machine} proposed the Set Cover Optimal Teaching (SCOT) algorithm for selecting demonstrations that provide the tightest constraints on a target reward function for a pure IRL learner. However, human learning is more multi-faceted and our prior work \cite{lee2021machine} tailored SCOT for humans by incorporating human learning techniques such as scaffolding. Our initial method of scaffolding via IRL did not yield significant learning gains, which we aim to improve in this work by incorporating counterfactuals based on the human's beliefs regarding the robot's reward function.


Finally, we note that there are other IRL methods commonly used in the literature such as Bayesian \cite{ramachandran2007bayesian} and MaxEnt \cite{ziebart2008maximum} IRL. Both were both used by Huang et al. \cite{huang2019enabling} as alternate models of how humans may recover the reward function underlying selected demonstrations. Though these models can incorporate the learner's beliefs (and subsequent counterfactuals) via beliefs initially sampled from a prior over the robot's reward function, such beliefs are not resampled with additional demonstrations in \cite{huang2019enabling}, making these IRL methods sensitive to the initial sampling and perhaps leading to slower convergence to the robot's reward function.

\textbf{Techniques for Human Teaching:} The following ideas from cognitive science inspire our approach for how a robot may convey its decision-making to a human learner.

\textit{Scaffolding:} This pedagogical strategy informs how a teacher may assist a learner in grasping a task outside the learner's current skill set, e.g. by reducing the task's degrees of freedom \cite{wood1976role}. We implement this by showing demonstrations that convey information on an increasing number of unique reward features. Following Reiser's \cite{reiser2004scaffolding} recommendation to reduce learning complexity through scaffolding additional structure, we also provide demonstrations that sequentially increase in both informativeness and difficulty.

\textit{Counterfactuals:} Literature on how humans explain to one another notes that ``explanations are contrastive---they are sought in response to particular counterfactual cases,'' and that it is critical that the learner's counterfactuals matches the ones intended by the teacher \cite{miller2019explanation}. In our work, we tailor demonstrations to the learner given their current belief and the counterfactuals they will likely consider.
Furthermore, Reiser \cite{reiser2004scaffolding} suggests that scaffolding should sometimes challenge the learner by inducing cognitive conflict. It is thus important to ensure that the robot correctly anticipates the human's likely counterfactual and provides a demonstration that \textit{differs} from that counterfactual to provide information.

\textit{Testing:} A learner's current abilities must be accurately assessed to provide the right assistance when scaffolding \cite{collins1988cognitive}. We test learners accordingly by asking them to predict the robot's optimal behavior in unseen instances of a domain. 
In our prior work \cite{lee2021machine}, we showed that informativeness of a demonstration during teaching could simply be inverted to measure the expected difficulty of correctly predicting it during testing. In this work, we propose to update the difficulty measure by explicitly conditioning on the learner's beliefs of the robot's reward function. 

\section{Technical background}
\label{sec:technical_background}

\textbf{Markov decision process:} The robot models its world as an instance (indexed by $i$) of a Markov decision process, $MDP_i$, comprised of sets of states $\mathcal{S}_i$ and actions $\mathcal{A}$, a transition function $T_i$, reward function $R$, discount factor $\gamma$, and initial state distribution $S^0_i$. We refer to a group of related MDP instances as a \textit{domain} (described below) and \(\mathcal{S} : \bigcup_i \mathcal{S}_i\) is the union over all of their states. An optimal trajectory $\xi^*$ is a sequence of $(s_i, a, s_i')$ tuples obtained by following the robot's optimal policy $\pi^*$. Following prior work \cite{abbeel2004apprenticeship}, $R = \mathbf{w^*}^\top \phi(s, a, s')$ is represented as a weighted linear combination of reward features.
Finally, we assume the human is aware of the full MDP apart from weights $\mathbf{w^*}$.

A domain is a group of MDPs that share $R, \mathcal{A},$ and $\gamma$ but differ in $T_i, \mathcal{S}_i,$ and $S^0_i$. For example, all MDPs in the delivery domain share the same $R$ even though they may contain different mud patches (Figs. \ref{fig:combined}a and \ref{fig:combined}b). Thus through IRL, all demonstrations within a domain will support inference over a common $\mathbf{w}^*$. We simplify the notation such that $\pi^*$ refers to any optimal policy within a domain, and $\xi^*$ refers to a demonstration (dropping the corresponding MDP).

\textbf{Machine teaching for policies:} Our objective to select informative demonstrations for conveying $\pi^*$ is captured by the machine teaching framework for policies \cite{lage2019exploring}. We aim to select a set of demonstrations $\mathcal{D}$ of size $n$ that maximizes the similarity $\rho$ between optimal policy $\pi^*$ and the policy $\hat{\pi}$ recovered using a computational model $\mathcal{M}$ (e.g., IRL) on $\mathcal{D}$
\begin{equation}
    \argmax_{\mathcal{D} \subset \Xi} \rho(\hat{\pi}(\mathcal{D}, \mathcal{M}), \pi^*) \;\; \mathrm{s.t.} \;\; |\mathcal{D}| = n
\end{equation}
where $\Xi$ is the set of all demonstrations of $\pi^*$ in a domain. Once $\mathbf{w}^*$ is approximated through IRL, this approach assumes that the learner is able to deduce $\pi^*$ by planning on the underlying MDP. Thus, the objective reduces to selecting demonstrations that are informative at conveying $\mathbf{w}^*$, which can be measured using behavior equivalence classes.

\textbf{Behavior equivalence class:} The \textit{behavior equivalence class} (BEC) of a demonstration is the region of reward functions under which the demonstration is still optimal.

For a reward function that is a weighted linear combination of features, the BEC of a demonstration $\xi$ of $\pi$ is defined as the intersection of half-spaces \cite{brown2019machine} formed by the standard IRL equation \cite{ng2000algorithms}
\begin{equation}
\textrm{BEC}(\xi|\pi) := \mathbf{w}^{\top}\left(\mu_{\pi}^{(s, a)}-\mu_{\pi}^{(s, b)}\right) \geq 0, \forall(s, a) \in \xi, b \in \mathcal{A}.
    \label{eq:BEC_demo}
\end{equation}
where $\mu_{\pi}^{(s, a)}=\mathbb{E} \left[ \sum_{t=0}^{\infty} \gamma^{t} \phi\left(s_{t}\right) \mid \pi, s_{0}=s, a_{0}=a \right]$ is the vector of reward feature counts accrued from taking the action $a$ in $s$, then following $\pi$ after. Any demonstration can be converted into a set of constraints on $\mathbf{w}$ using (\ref{eq:BEC_demo}).

\begin{figure}[!t]
\centering
\includegraphics[width=\columnwidth]{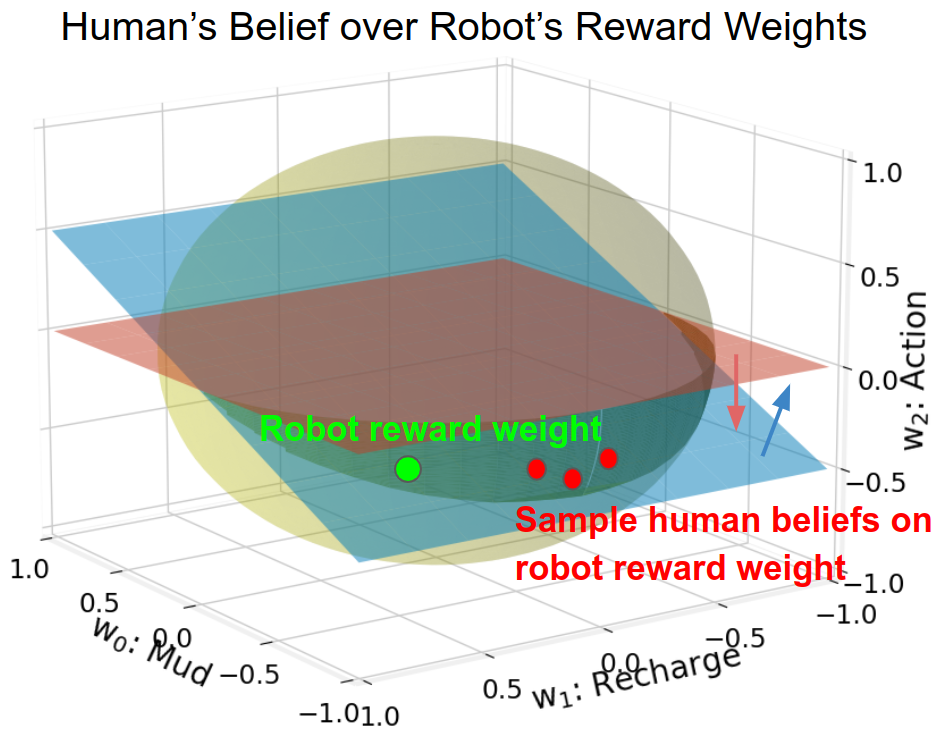}
\caption{Standard IRL compares a demonstration against counterfactuals to generate constraints on the possible underlying reward function. The red/blue half-space constraints on the hollow sphere of possible reward weights correspond to the information provided by the demonstration in Fig. \ref{fig:combined}a, and models the human's subsequent belief over the robot's reward function. The three samples of the human's belief (shown in red) would all give rise to the human counterfactual in Fig. \ref{fig:combined}b, as each assigns a high cost to mud.}
\label{fig:human_model}
\vspace{-3mm}
\end{figure}

Consider again the delivery domain, which has binary reward features $\mathbf{\phi} =$ [\emph{traversed mud}, \emph{battery recharged}, \emph{action taken}], $\mathbf{w^*} \propto [-3, 3.5, -1]$. In practice, we require $||\mathbf{w^*}||_2 = 1$ to bypass both the scale invariance of IRL and the degenerate all-zero reward function. With no information, we assume that the potential reward weights in the human's mind uniformly span the full surface of the $n-1$ sphere due to the $L^2$ norm constraint on $\mathbf{w^*}$, where $n$ is the number of domain features. The demonstration in Fig. \ref{fig:combined}a yields the constraints in Fig. \ref{fig:human_model}, which has left only a sliver of the 2-sphere surface which contains the robot's true reward weights. The red plane in Fig. \ref{fig:human_model} intuitively indicates the constraint that $w^*_2 \leq 0$, since no unnecessary actions were taken in delivering the package, and jointly with the blue plane indicates the constraint that $w^*_0 \leq 0$ and $w^*_0 \leq 2w^*_2$, since two actions were taken to detour around the mud. Importantly, the surface area of the $n-1$ sphere that remains after incorporating a demonstration's constraints can be used as a measure of its informativeness. The smaller the area, the fewer viable reward weights remain, and the more informative the demonstration.

\section{Proposed Techniques for Teaching Humans}

We make two observations regarding the BEC. First, (\ref{eq:BEC_demo}) captures the key idea that IRL depends not only on the robot's optimal trajectory but also on the suboptimal counterfactual trajectories that are considered, represented by $\mu_{\pi}^{(s, a)}$ and $\mu_{\pi}^{(s, b)}$ respectively. Second, BEC($\xi|\pi$) could be used to model the human's belief over the robot's possible reward weights after having seen $\xi$. We now build on these concepts to leverage a human model and select demonstrations that account for human counterfactuals\footnote{Code for the methods, domains, and hyper-parameters can be found \href{https://github.com/SUCCESS-MURI/counterfactual_human_IRL}{here}.}.

\textbf{Counterfactual Scaffolding:} A demonstration's ability to reveal the underlying reward function via IRL hinges on the counterfactuals considered. However, many counterfactuals proposed by IRL can seem nonsensical to humans as they fail to consider the human's beliefs. Instead, IRL generates counterfactuals in the following way—at each state $s$ along the robot's optimal trajectory, it first takes a potentially suboptimal action $b$ before following the optimal policy afterward (\ref{eq:BEC_demo}). This process generates the two sample counterfactuals in Fig. \ref{fig:combined}c, which do not correspond to the human counterfactual in Fig. \ref{fig:combined}b. While such one-action deviations from the optimal trajectory are computationally sensible and efficient (as multi-action deviations often only yield redundant constraints), these are unlikely to be the counterfactuals on the human's mind for a number of reasons.

First, humans are unlikely to methodically go through each state of the robot's trajectory and consider all alternative actions. Instead, humans naturally incline toward a few causes and a few counterfactuals for explanation \cite{miller2019explanation}. This can lead IRL to overvalue the informativeness of a demonstration if counterfactuals beyond those on the human's mind are considered. Second, IRL counterfactuals are generated by ``perturbing'' the demonstration directly (by taking a suboptimal action) and may not be consistent with any reward function (e.g. no reward function in the delivery domain would first avoid mud, then later go through mud like one of the counterfactuals in Fig. \ref{fig:combined}c). While humans may consider a reward function that differs from the robot's, their counterfactuals are likely to be consistent with that ``perturbed'' reward function (e.g. avoiding the mud both ways in Fig. \ref{fig:combined}b). This can lead IRL to also undervalue a demonstration's informativeness if the human's counterfactuals are not considered.

In selecting effective explanations, we posit that you must not only consider the learner's learning model (i.e. IRL) but also their beliefs and subsequently what counterfactuals they would consider. We thus extend our previous work \cite{lee2021machine} to evaluate a demonstration's informativeness based on counterfactuals generated via potential reward functions on the human's mind as opposed to counterfactuals generated via one-action deviations, which can be expressed as
\begin{equation}
\textrm{BEC}_{CF}(\xi|\pi, \pi_{\mathbf{w}}) := \mathbf{w}^{\top}\left(\mu_{\pi}^{s}-\mu_{\pi_{\mathbf{w}}}^{s}\right) \geq 0, s = \xi(0),
    \label{eq:BEC_cf}
\end{equation}
where $\mu^{s}_\pi= \mathbb{E}\left[\sum_{t=0}^{\infty} \gamma^{t} \phi\left(s_{t}\right) \mid \pi, s_0 = s \right]$ is the vector of reward feature counts accrued from starting in $s$ and following $\pi$ after (note that $\pi_w$ is the optimal policy under reward weight $w$) and $\xi(0)$ is the first state of $\xi$. We can now scaffold by showing demonstrations of increasing informativeness.

To account for human beliefs and counterfactuals when evaluating the informativeness of potential demonstrations, we do the following. First, we instantiate a prior model of the human's beliefs over the robot's reward weights $\mathbf{w}^*$ as $B(\mathbf{w}^*)$. This model could be the full $n-1$ sphere if the human has no prior knowledge, or it may be a partial sphere due to prior knowledge (e.g. action weight is negative). Then we sample $m$ weights from $B(\mathbf{w}^*)$. Each weight $\mathbf{w}$ represents a particular belief that the human could have over the robot's reward function. For every possible robot demonstration in a domain, and for each of the $m$ weights, we simulate what the ``human'' counterfactual to each demonstration would be if the human had this reward weight in mind and generate the corresponding constraints using (\ref{eq:BEC_cf}). For each possible demonstration by the robot, we consolidate the corresponding $m$ human counterfactuals by taking a union of all corresponding constraints. Finally, we select the demonstration that differs strongly from the human's expectations and subsequently maximizes knowledge gain, i.e. select the demonstration that maximizes the ratio between $B(\mathbf{w}^*)$ before and after the human sees this demonstration. Once we have shown the selected demonstration and updated $B(\mathbf{w}^*)$, we select the next demonstration to show by sampling $m$ weights from the updated $B(\mathbf{w}^*)$ and repeating the steps above. This method is summarized in Alg. \ref{alg:counterfactual_scaffolding}, where $\mathbf{\hat{N}}[\cdot]$ denotes the operation of extracting unit normal vectors corresponding to a set of half-space constraints, and $\setminus$ denotes set subtraction.

\begin{algorithm}[t]
\caption{Counterfactual Machine Teaching for Humans}\label{alg:counterfactual_scaffolding}
\begin{algorithmic}[1]
\Require $\pi^*$: robot policy, $\Xi$: all possible demonstrations of $\pi^*$ in a domain, $m$: number of beliefs to sample, $B(\textbf{w}^*)$: human prior over robot reward weights

\State $\mathrm{knowldgGn} = \infty$ \Comment{Knowledge gain}
\State $\mathcal{D} = [\:]$

\While{$\mathrm{knowldgGn} \neq 0$}
\State $B_{dict} = \varnothing$
\State // Sample human beliefs on $\textbf{w}^*$
\State $\textbf{W} = \mathrm{sample(\textit{m}, B(\textbf{w}^*))}$

\State // Obtain constraints yielded by each possible demonstration, conditioned on the sampled human beliefs
\For{$\xi \in \Xi$}
\State $\mathcal{C} = \varnothing$
\For{$\mathbf{w} \in \textbf{W}$}
\State // Constraints given ``human'' counterfactual
\State $C = C \cup \mathbf{\hat{N}}\mathrm{[BEC}_{CF}(\xi | \pi^*, \pi_\mathbf{w})]$ \Comment{Eq. \ref{eq:BEC_cf}}
\EndFor
\State // Store updated belief given this demonstration
\State $B_{dict}[\xi] = C \cup \mathbf{\hat{N}}[B(\textbf{w}^*))]$
\EndFor

\State // Select trajectory that maximizes knowledge gain
\State{\small $\xi^* = \argmax\limits_{\xi \in \Xi} \mathrm{BECArea}(B(\mathbf{w}^*)) \, / \, \mathrm{BECArea}(B_{dict}[\xi]) $}
\State {\small $\mathrm{knowldgGn} = \mathrm{BECArea}(B(\mathbf{w}^*)) \, /\, \mathrm{BECArea}(B_{dict}[\xi^*]) $}

\If{$\mathrm{knowldgGn} \neq 0$}
\State $\mathcal{D}.\mathrm{append}(\xi^*)$
\State $\Xi = \Xi \setminus \xi^*$
\State $B(\mathbf{w}^*) = B_{dict}[\xi]$ \Comment{Update human belief}
\EndIf
\EndWhile

\State \textbf{return} $\mathcal{D}$\Comment{Final demonstration set to show human}

\end{algorithmic}
\end{algorithm}

\begin{algorithm}[t]
\caption{Measuring Test Difficulty of a Demonstration}\label{alg:testing}
\begin{algorithmic}[1]
\Require $\pi^*$: robot policy, $\xi$: demo, $m$: number of beliefs to sample, $B(\textbf{w}^*)$: human prior over robot reward weights

\State $\mathrm{BEC}^\prime(\xi | \pi^*) = \varnothing$

\State // Sample possible beliefs on $\textbf{w}^*$
\State $\textbf{W} = \mathrm{sample(\textit{m}, B(\textbf{w}^*))}$

\State // Obtain constraints yielded by each possible demonstration, using both standard IRL and human counterfactuals
\State $\mathrm{BEC}^\prime(\xi | \pi^*) = \mathrm{BEC}^\prime(\xi | \pi^*) \cup \mathbf{\hat{N}}\mathrm{[BEC}(\xi | \pi^*)]$
\For{$\mathbf{w} \in \mathbf{W}$}
\State $\mathrm{BEC}^\prime(\xi | \pi^*) = \mathrm{BEC}^\prime(\xi | \pi^*) \cup \mathbf{\hat{N}}[\mathrm{BEC}_{CF}(\xi | \pi^*, \pi_{\mathbf{w}})]$
\EndFor
\State // The overlap is inversely correlated to difficulty
\State{$\mathrm{difficulty = 1 / measureOverlap}(B(\textbf{w}^*), \mathrm{BEC}^\prime(\xi | \pi^*))$}
\State \textbf{return} $\mathrm{difficulty}$
\end{algorithmic}
\end{algorithm}
\textbf{Feature Scaffolding:} A standard scaffolding technique suggested by Wood et al. \cite{wood1976role} is to reduce the degrees of freedom of the problem. Accordingly, one could initially show demonstrations that limit the number of reward features over which information is conveyed. In the delivery domain, one could show demonstrations that convey information on the mud and action weights first, then on battery recharge and action weights, then on mud, battery, and action weights to show potentially nuanced tradeoffs\footnote{Because solutions of IRL are scale invariant, we must show at least two features at a time relative to one another (e.g. how many actions are you willing to take to avoid mud?).}. Any feature over which information should not be conveyed is ``masked''.

To scaffold three features, we first determine the order that the features will be masked. For all of the possible robot demonstrations in a domain, we obtain all possible constraints that could be generated using (\ref{eq:BEC_demo}). We mask features in order from the one with fewest number of nonzero entries across all of the constraints\footnote{The sample constraint $\textbf{w}^\top [2, 0, -5] \geq 0$ has nonzero entries for the first and third features.} to the one that has the greatest number, with the rationale that a high number of nonzero entries are often indicative of a good reference feature useful for comparisons early on (e.g. the two-action deviation denotes mud as costly in Fig. \ref{fig:combined}a). Once the masking order has been decided, we remove any demonstrations that convey information about the first masked feature from consideration (i.e. any demonstrations that conveys constraints in which the entry for a masked feature is nonzero). From this reduced set of demonstrations, we run counterfactual scaffolding as described in the previous subsection until there are no more demonstrations that can provide additional knowledge gain. We then remove any demonstrations that convey information about the second masked feature from consideration and run counterfactual scaffolding until there are no more demonstrations that can provide additional knowledge gain. We repeat for the third masked feature. Finally, we consider all possible demonstrations in the domain and run counterfactual scaffolding until there are no more demonstrations that can provide additional knowledge gain. Generalizing to higher dimensions, we can scaffold $k$ features by showing demonstrations that iteratively mask $k-2$, $k-3$, ..., $0$ features and showing combinations of two, three, ..., all features respectively.

\textbf{Testing:} The area of a demonstration's BEC intuitively correlates inversely with its informativeness during teaching as smaller areas indicate less uncertainty regarding $\mathbf{w}^*$. Our prior work \cite{lee2021machine} showed that a demonstration's BEC area may also be inverted to measure the difficulty of correctly predicting the demonstration as a test if the human has not seen it before (i.e. smaller BEC area indicates a difficult test).

\begin{figure}[t]
\centering
\includegraphics[width=0.8\columnwidth]{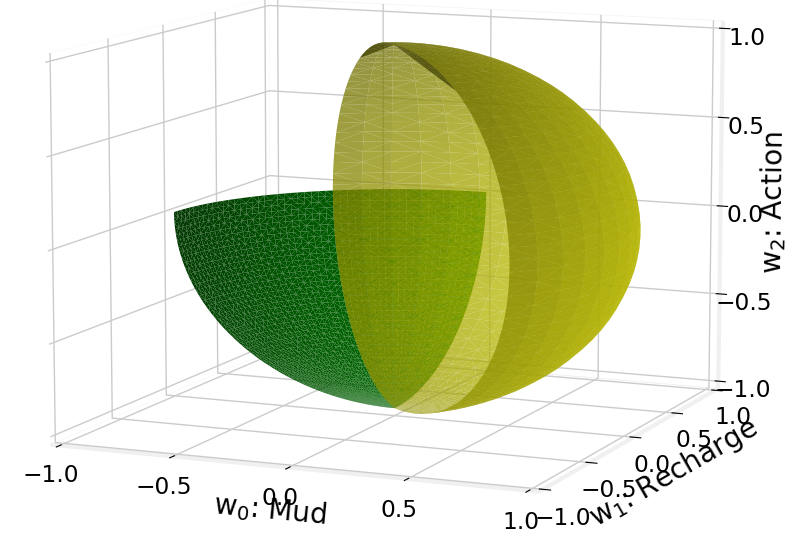} \caption{There are many reward weights BEC($\xi|\pi^*$) (yellow) that will generate the demonstration $\xi$. However, only a portion overlaps with the weights currently on the human's mind $B(\mathbf{w}^*)$ (green), making it difficult for the human to correctly predict $\xi$ during testing. }
\label{fig:overlap}
\vspace{-4mm}
\end{figure}

However, the learner's belief over the robot's reward function also likely plays a role. We hypothesize that the overlap between BEC($\xi|\pi^*$) and $B(\mathbf{w}^*)$ better captures the difficulty of a demonstration $\xi$ as a test for a human. This overlap intuitively represents the number of candidate reward function in the human's mind that would generate the correct behavior. As seen in Fig. \ref{fig:overlap}, a demonstration may have an intrinsically large BEC area but not overlap much with the human's belief and therefore be a difficult test.

To estimate the expected difficulty of correctly predicting $\xi$ as a test, we first obtain BEC($\xi|\pi^*$) using (\ref{eq:BEC_demo}). Noting that one-action deviation does not always consider all reasonable counterfactual trajectories, we take a union over constraints that define BEC($\xi|\pi^*$) and constraints obtained from counterfactual scaffolding using $m$ models sampled from $B(\mathbf{w}^*)$ using (\ref{eq:BEC_cf}). These combined constraints for each demonstration will give a better estimate of the set of all weights that yield the correct demonstration, denoted by BEC$^\prime$($\xi|\pi^*$). Finally, to measure the difficulty of a demonstration $\xi$ as a test for this human, we simply take the overlap between $B(\mathbf{w}^*)$ and BEC$^\prime$($\xi|\pi^*$). The smaller the overlap, fewer of the reward functions in the human's mind will generate the correct demonstration, and the harder the test. This method is summarized in Alg. \ref{alg:testing}.

\section{User Study}

\begin{figure*}
\centering
\includegraphics[width=0.9\textwidth]{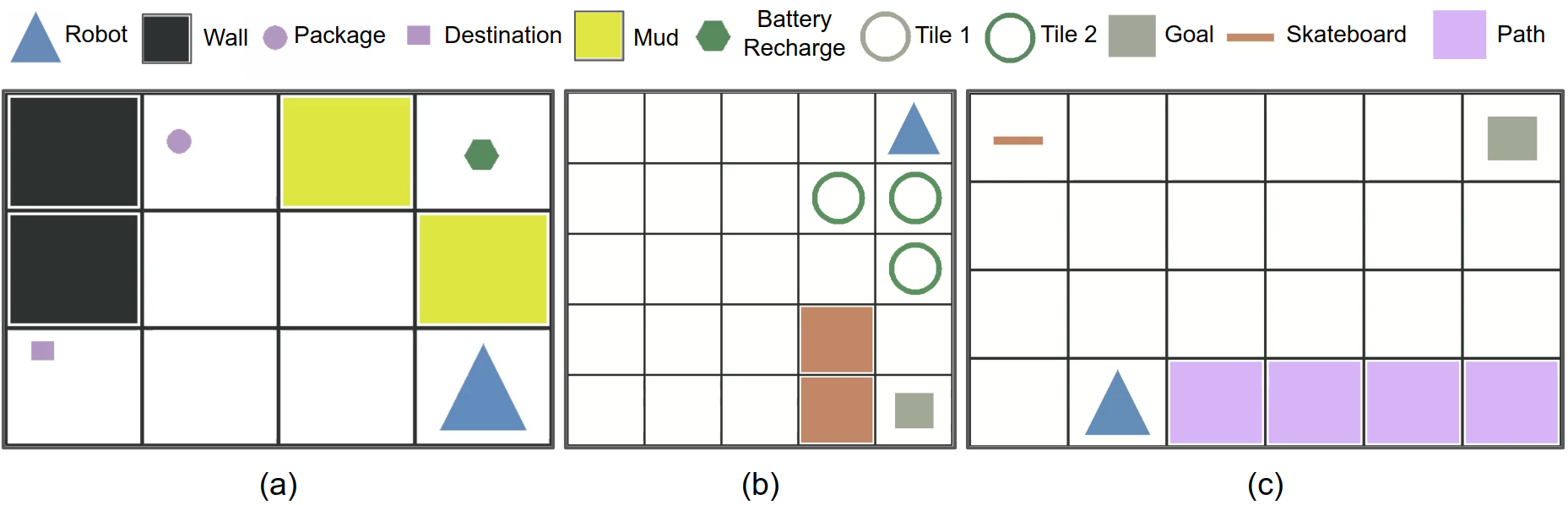}
\caption{Three domains were designed for the user study, each with a different set of reward weights to infer from demonstrations: \textbf{(a)} delivery  \textbf{(b)} tiles \textbf{(c)} skateboard. The semantics of the various objects were hidden using abstract geometric shapes and colors.}
\label{fig:domains3}
\vspace{-3mm}
\end{figure*}

We ran an online user study\footnote{Code for the user study, data, and analyses can be found \href{https://github.com/SUCCESS-MURI/counterfactual_human_IRL_study}{here}.} exploring whether  demonstrations selected using counterfactual and feature scaffolding improves a human's understanding of a robot's policy. The study involved participants watching robot demonstrations in three deterministic domains and predicting the robot's behavior in new test environments. Each domain consisted of one shared action reward feature (that helped penalize each action), and two unique reward features as follows (see Fig. \ref{fig:domains3}).

\textbf{Delivery domain.} The robot is penalized for moving out of mud and rewarded for recharging. Five demonstrations were shown in this domain.

\textbf{Tiles domain.} The robot is penalized differently for traversing the two differently shaped tiles. Five demonstrations were shown in this domain.

\textbf{Skateboard domain.} The robot is penalized less per action if it has either picked up a skateboard (i.e. riding is less costly than walking) or is traversing through a designated path. Seven demonstrations were shown in this domain.

The participants were explicitly informed of each domain's reward features, but had to infer the respective reward weights by watching demonstrations. The demonstrations they were provided were determined by the one of four between-subjects conditions they were assigned.

The between-subjects variables were \emph{BEC scaffolding} (counterfactual and baseline), and \emph{feature scaffolding} (yes and no). Baseline scaffolding followed the method proposed by our prior work \cite{lee2021machine}, using one-action deviations to generate counterfactuals and selecting demonstrations that iteratively decreased in BEC area. And no feature scaffolding did not hold out demonstrations of iteratively mask features as feature scaffolding did.\footnote{For the `counterfactual' and `no feature scaffolding' condition, demonstrations that provided 70\% of maximal knowledge gain were selected for the delivery and skateboard domains so that all between-subjects conditions provided the same number of demonstrations (else this condition would have shown fewer demonstrations than the others). We note that demonstrations from all conditions yielded the same final BEC area for each domain, theoretically providing the same amount of information in the end.} We conservatively modeled $B(\mathbf{w}^*)$ prior to any demonstrations having being shown as knowing that the action weight is negative (assuming a human bias for efficiency). All four between-subjects conditions optimized visual similarity amongst consecutive demonstrations and visual simplicity within demonstrations, as suggested by our prior work \cite{lee2021machine}. Thus, for any given set of equally informative demonstrations, the one that looked most similar to the previously shown demonstration (e.g. location of mud patches) and also had the fewest visual clutter (e.g. number of mud patches) was selected to be shown next.

There were also two within-subject variables: \emph{domain} (delivery, tiles, skateboard) and \emph{test difficulty} (high, medium, and low). The tests were pulled from three representative sets of demonstrations that had high, medium, and low overlap between $B(\mathbf{w}^*)$ and BEC$^\prime$($\xi|\pi^*$) for the low, medium, and high difficulty test conditions. Specifically, the overlap for every possible demonstration in a domain were ordered from high to low, grouped into five clusters using K-means, and high, medium, and low overlap demonstrations were taken from the 1st, 3rd, and 5th cluster respectively. We conservatively modeled $B(\mathbf{w}^*)$ after having watched all of the teaching demonstrations as knowing the correct sign (i.e. positive or negative) of each of the reward weights.

The user study itself primarily consisted of three trials, with each trial comprising a teaching portion and a testing portion in a unique domain. During teaching, participants were first explicitly informed of the reward features of the domain. Then they inferred the corresponding reward weights by watching demonstrations\footnote{To mitigate effects of limited memory, participants were allowed to rewatch demonstrations at any point before moving on to the testing portion.} and provided subjective observations regarding the demonstrations (M2-M3, see the following paragraph). For testing, participants were tasked with predicting the optimal trajectory in six unseen test environments (a random order of two high, medium, and low difficulty environments each, M1) and rating their confidence in their responses (M4). The following measures (M1-M4) were used to test the hypotheses below (H1-H4).

\noindent \textbf{M1. Optimal response:} Participants were assigned a binary score depending on the optimality of their test trajectory.

\noindent \textbf{M2. Informativeness rating:} ``How informative were these demonstrations in understanding the best strategy [robot's policy] in this game?'', answered with a 5-point Likert scale

\noindent \textbf{M3. Mental effort rating:} ``How much mental effort was required to understand the best strategy [robot's policy] in this game?'', answered with a 5-point Likert scale

\noindent \textbf{M4. Confidence rating:} ``How confident are you that you [performed the task optimally in this unseen test environment]?'', answered with a 5-point Likert scale

\textbf{H1}: The overlap between a human's belief over the weights $B(\mathbf{w}^*)$ and the BEC of a demonstration BEC$^\prime$($\xi|\pi^*$) correlates inversely to the difficulty of predicting it during testing and correlates directly to their prediction confidence.

\textbf{H2}: Demonstrations selected with counterfactual scaffolding will result in higher perceived informativeness during teaching and better participant test performance over those selected with baseline scaffolding \cite{lee2021machine}.

\textbf{H3}: Demonstrations selected with feature scaffolding will result in lower mental effort during teaching and better participant test performance over those selected without.

\textbf{H4}: Demonstrations selected with counterfactual scaffolding and feature scaffolding will result in the highest perceived informativeness of teaching demonstrations, lowest mental effort, and best participant test performance.

\section{Results and Discussion}

We collected data from 216 participants using Prolific. Participants were roughly 67\% male, 32\% female, 1\% non-binary and ages varied from 18 to 69 (M = 28.39, SD = 9.48). 54 participants were randomly assigned to each of the four between-subjects conditions and the order in which the domains were shown was fully counterbalanced. We removed data from 4 participants whose aggregate test performance (compared to the optimal answer)\footnote{Calculated by averaging each participant's 18 test responses (i.e. six tests in three domains) into a percentage of tests that the participant got correct.} or many individual test performances (compared to other participants)\footnote{Calculated by comparing an individual's test performances against other participants. The number of times a participant's reward for a test trajectory was 3 standard deviations below the mean reward was compared.} were 3 standard deviations below their respective means as outliers.


The three domains varied in the difficulties of their optimal policies. We calculated a mean-rating (k = 3), 2-way mixed effects, consistency-based intraclass coefficient (ICC) to see how the performance of each participant varied across domains \cite{koo2016guideline}. Given an ICC of 0.32 that indicates significant variance ($p < .001$), we average the performance of every participant across domains and provide findings that may represent a range of domains and difficulties.

\textbf{H1:} A one-way repeated measures ANOVA on percentage of optimal responses revealed a statistically significant difference across test difficulty ($F(2,422) = 289.78, p < .001$). Post-hoc pairwise Tukey analyses confirmed significant differences between high ($M = 0.40$), medium ($M = 0.71$), and low ($M = 0.86$) test difficulties ($p < .001$ in all cases).

Spearman's rank-order correlation revealed that confidence inversely correlated significantly with test difficulty ($r_s = -.36, p < .001, N = 636$).

\textit{The data support H1} that the overlap between $B(\mathbf{w}^*)$ and BEC$^\prime$($\xi|\pi^*$) captures a demonstration's difficulty for testing and confirm \textit{test difficulty} as a valid within-subjects variable.

\begin{figure}[t]
  \centering
  \includegraphics[width=0.85\columnwidth]{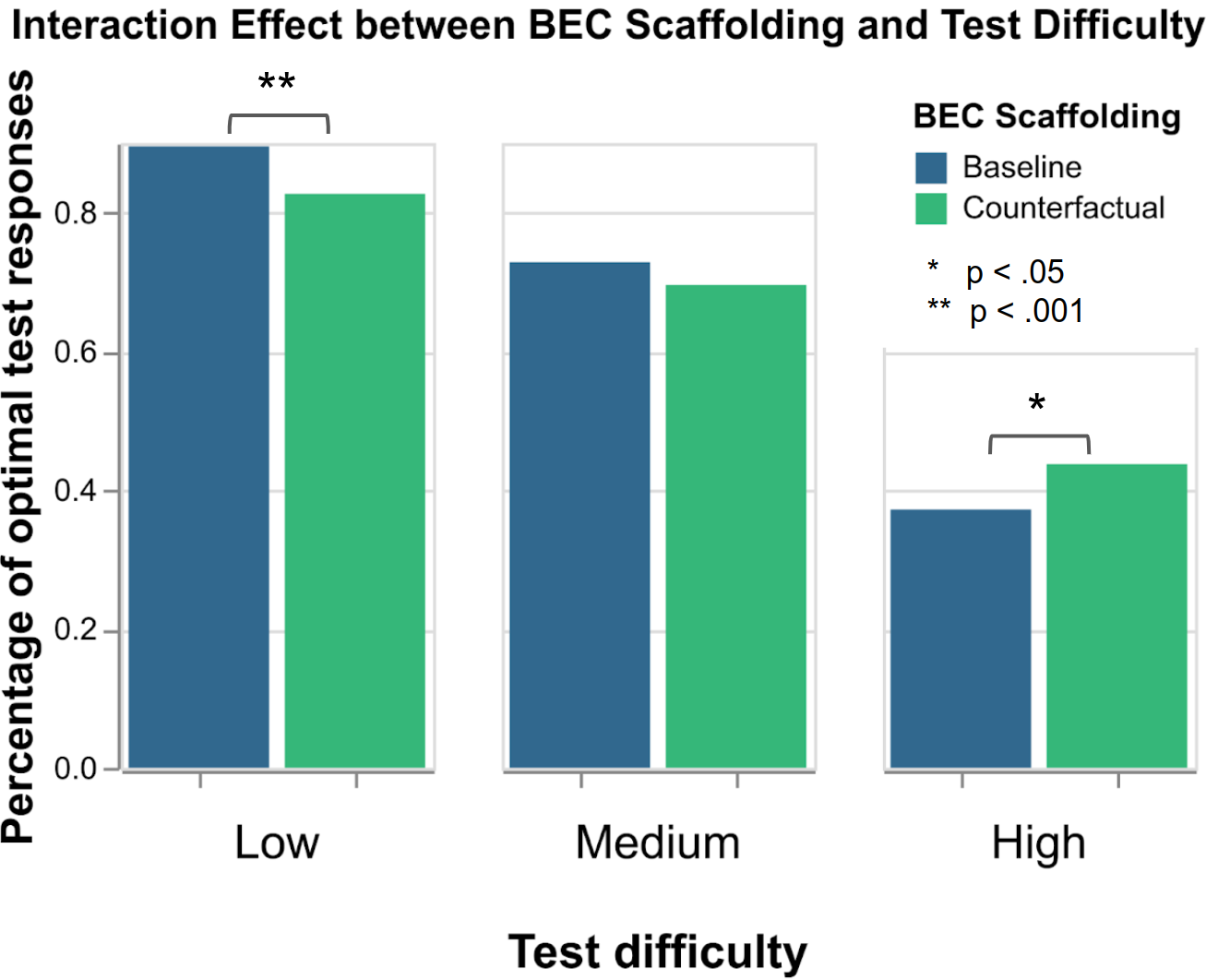}
  \captionof{figure}{While baseline scaffolding significantly increases performance on low difficulty tests over counterfactual scaffolding, the effect is reversed for high difficulty tests.}
  \label{fig:H2_interaction}
  \vspace{-5mm}
\end{figure}

\textbf{H2:} A two-way mixed ANOVA revealed a significant interaction between counterfactual scaffolding and test difficulty for percentage of optimal responses ($F(2,420) = 6.56, p = .002$). Tukey analyses revealed that for low difficulty tests ($p = .002$), no counterfactual scaffolding ($M = 0.90$) significantly improved performance over counterfactual scaffolding ($M = 0.83$). However, the relationship was reversed for high difficulty tests ($p = .048$) with counterfactual scaffolding ($M = 0.44$) outperforming no counterfactual scaffolding ($M = 0.37$), as shown in Fig. \ref{fig:H2_interaction}. Surprisingly, a significant effect was not revealed for counterfactual scaffolding by a two-way mixed ANOVA ($F(1,210) = 0.47, p = .49$).

Ratings for mental effort was found by a Mann-Whitney U test to be significantly higher for counterfactual scaffolding ($U(N_{baseline} = 108, N_{counterfactual} = 104) = 4690.0, p = .03$). The two counterfactual scaffolding conditions did not differ significantly in informativeness ratings ($p = .08$).

As exploratory measures, we also recorded the average number of times a participant watched each demonstration and the time taken for a participant to provide a test demonstration and rate their confidence. Interestingly, Tukey analyses revealed that counterfactual scaffolding significantly increased the average number of times a teaching demonstration was watched $(M = 1.23)$ over baseline scaffolding $(M = 1.15, p = .02)$ and also significantly increased the time taken to complete a test $(M = 2.95\textrm{ sec})$ over baseline scaffolding $(M = 2.49 \textrm{ sec}, p = .01)$.

\textit{The data partially support H2.} Counterfactual scaffolding fails to outperform baseline scaffolding in test performance. However, counterfactual scaffolding appears to improve test performance on high difficulty tests at the cost of increased mental effort (as indicated by both objective and subjective measures). As previously mentioned, Reiser \cite{reiser2004scaffolding} suggests that scaffolding should sometimes challenge and engage the learner by inducing cognitive conflict. Indeed counterfactual scaffolding explicitly selects demonstrations that would not be anticipated by the learner and requires the learner to reconcile the gap by updating their belief.

Finally, counterfactual scaffolding performs worse than baseline scaffolding for low difficulty tests. In our counterfactual scaffolding method, we always took highly informative demonstrations given the user's current belief. However, a person's learning ability is likely more context dependent (e.g. on their prior knowledge, current stage of learning, etc) and learning rate should be more personalized in the future.

\textbf{H3:} A two-way mixed ANOVA revealed that feature scaffolding had no significant effect on percentage of optimal responses ($F(1,210) = 1.79, p = .18$), and no interaction between feature scaffolding and test difficulty ($F(2,420) = 1.72, p = .18$). Mann-Whitney U tests found that feature scaffolding did not impact ratings on informativeness ($p = .81$) nor mental effort ($p = .14$).

\textit{The data does not support H3.} We did not observe any effect for feature scaffolding. The domains each only had three reward features, which perhaps were already too few to significantly benefit from feature scaffolding.

\textbf{H4:} A two-way mixed ANOVA revealed a significant interaction effect between the four possible between-subjects conditions and test difficulty on percentage of optimal responses ($F(6,416) = 4.40, p < .001$). Tukey analyses showed that counterfactual scaffolding with no feature scaffolding ($M = 0.75$) was significantly outperformed by baseline scaffolding with ($M = 0.88$) and without ($M = 0.91$) feature scaffolding, and also by counterfactual scaffolding with feature scaffolding ($M = 0.91$) for low test difficulty. The conditions did not significantly affect test performance ($F(3,208) = 1.98, p = .12$).

Mann-Whitney U tests revealed that counterfactual scaffolding with feature scaffolding (the proposed method) required the most mental effort (Mdn = 3) over baseline scaffolding with (Mdn = 2, $p < .001$) and without (Mdn = 2, $p = .007$) feature scaffolding, and also over counterfactual scaffolding without feature scaffolding (Mdn = 2, $p = .005$).

\textit{The data does not support H4.} The observations that counterfactual scaffolding can decrease performance on low difficulty tests and requires more mental effort corresponds with the analysis for H2. Though not statistically significant, once again, perhaps the increase in mental effort is correlated with the increase in \textit{overall} and \textit{high test difficulty performance} of counterfactual scaffolding with feature scaffolding ($M = 0.68, M = 0.45$ respectively), over baseline scaffolding with ($M = 0.66, M = 0.35$) and without ($M = 0.67, M = 0.39$) feature scaffolding and counterfactual scaffolding without feature scaffolding ($M = 0.63, M = 0.42$).

\section{Conclusion}
Our ability to collaborate well with robots is contingent on us understanding their decision making. We thus provided a method for robots to convey their decision making to humans through demonstrations, leveraging the key insight that an informative demonstration is intuitively one that differs strongly from the human's expectations (i.e. counterfactuals). We also proposed a new measure for quantifying the difficulty for a human to predict an unseen instance of a robot's behavior as a test. A user study confirmed our test difficulty measure correlates with human test performance and confidence. The study also interestingly showed that our proposed demonstrations decreased human performance on easy tests but increased performance on difficult tests. While our proposed demonstrations were ultimately informative, perhaps the first few differed too significantly from the human's counterfactuals and were difficult to grasp. Effective pedagogy often interweaves teaching and testing, e.g. by adapting subsequent lessons to the learner's test performance. We leave this as an exciting direction for future work.

\bibliographystyle{IEEEtran}
\bibliography{root}

\end{document}